\documentclass[sigconf]{acmart}

\copyrightyear{2017}
\acmYear{2017}
\setcopyright{acmlicensed}
\acmConference{CIKM'17 }{November 6--10, 2017}{Singapore, Singapore}
\acmPrice{15.00}
\acmDOI{10.1145/3132847.3132878}
 
\acmISBN{978-1-4503-4918-5/17/11}

\usepackage{eqnarray,amsmath}
\usepackage{graphicx}
\usepackage{csquotes}
\usepackage[linesnumbered]{algorithm2e}
\usepackage{algpseudocode}
\usepackage{color}
\usepackage{url}
\usepackage{graphicx}
\usepackage{tabularx}
\usepackage{multirow,booktabs}
\usepackage{hhline}
\usepackage{arydshln}
\usepackage{mathtools}
\usepackage{paralist}
\usepackage{enumitem}
\usepackage{setspace}
\usepackage{caption}

\DeclarePairedDelimiterX{\norm}[1]{\lVert}{\rVert}{#1}
\setlist{nosep}
\usepackage{microtype}

\SetAlFnt{\small\sffamily}
\usepackage{booktabs} 
\usepackage{subcaption}

\newcommand{\mypar}[1]{\medskip\pagebreak[3]\noindent\textbf{#1}~}

\definecolor{darkblue}{rgb}{0.0,0.0,0.5}

\renewcommand{\theenumii}{\theenumi.\arabic{enumii}.}

\newcolumntype{C}{>{\centering\arraybackslash}X}

\begin{document}

\title{Words are Malleable: Computing Semantic Shifts in Political and Media Discourse}


\author{Hosein Azarbonyad}
\affiliation{%
  \institution{University of Amsterdam}
}
\email{h.azarbonyad@uva.nl}

\author{Mostafa Dehghani}
\affiliation{%
  \institution{University of Amsterdam}
}
\email{dehghani@uva.nl}

\author{Kaspar Beelen}
\affiliation{%
  \institution{University of Amsterdam}
}
\email{k.beelen@uva.nl}

\author{Alexandra Arkut}
\affiliation{%
  \institution{University of Amsterdam}
}
\email{aeearkut@gmail.com}

\author{Maarten Marx}
\affiliation{%
  \institution{University of Amsterdam}
}
\email{maartenmarx@uva.nl}

\author{Jaap Kamps}
\affiliation{%
  \institution{University of Amsterdam}
}
\email{kamps@uva.nl}

\begin{abstract}
Recently, researchers started to pay attention to the  detection of  temporal shifts  in the meaning of words. 
However, most (if not all) of these approaches restricted their efforts to uncovering change over time, thus neglecting other valuable dimensions such as social or political variability.
We propose an approach  for detecting semantic shifts   between different \emph{viewpoints}---broadly defined as a set  of texts that share a specific metadata feature, which can be a time-period, but also a social entity such as a political party. 
%
For each viewpoint, we learn a semantic space in which each word is represented as a low dimensional neural embedded vector.
The challenge is to compare the meaning  of a word in one space to its meaning in another space and measure the size of the semantic shifts. 
We compare the effectiveness of a measure based on  optimal transformations between the two spaces with a measure based on the similarity of the neighbors of the word in the respective spaces. Our experiments demonstrate that the  combination of these two performs best.
We show that the semantic shifts not only occur over time, but also along different viewpoints in a short period of time. For evaluation, we demonstrate how this approach  captures meaningful semantic shifts and can help improve  other tasks such as the contrastive viewpoint summarization and ideology detection (measured as classification accuracy) in political texts. 
We also show that the two laws of semantic change  which were empirically shown to hold for temporal shifts also hold for shifts across viewpoints. These laws state that frequent words are less likely to  shift meaning  while  words with many senses are more likely to do so.
\end{abstract}


\maketitle


\section{Introduction}
\label{introduction}

Words are always `under construction', their meaning is unstable and malleable \cite{wittgenstein1967, skinner1969, triandafyllidou2003, Lansdall2017}.
Semantic fluctuations can result from a concept's `essentially contested' nature. 
"What does democracy mean?"  or "what values are democratic?". The answer changes
according to the ideological perspective or \emph{viewpoint} \cite{gallie1955} of the person uttering the term. 
Equally important is the influence of historic events. The understanding of `terrorism', for example, has significantly changed as a result of the 9/11 attacks \cite{bleich2016, reese2009}.
Currently, only a few studies have attempted to compute the `malleability of meaning' and monitor semantic shifts \cite{Jatowt:2014:FAS:2740769.2740809, Kenter2015, Kulkarni2015, Hamilton2016}. Most (if not all) of these approaches have focused their efforts to uncovering change over time. However, there are other valuable dimensions that can cause semantic shifts such as social or political variability. 

\begin{figure*}
\centering
\begin{subfigure}{.45\textwidth}
  \centering
   \caption{Shift over both time and political context dimensions}
  \includegraphics[width=\linewidth]{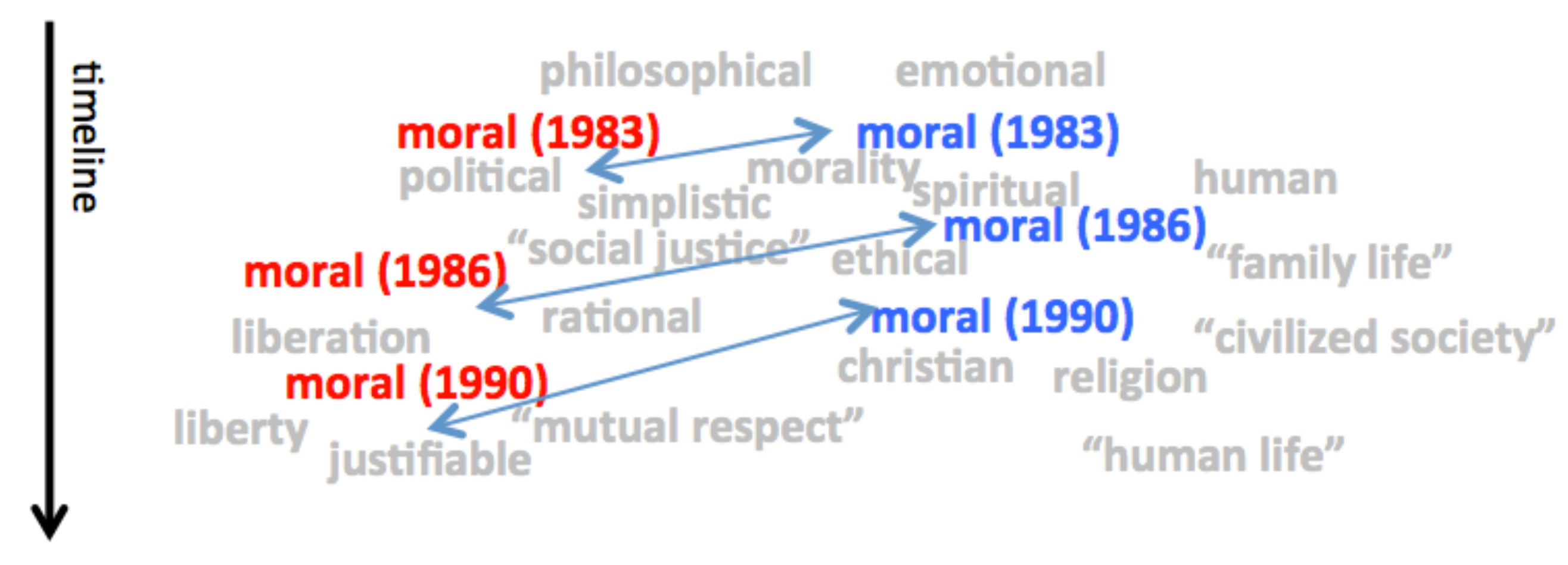}
  \label{fig:sub1}
\end{subfigure}%
\begin{subfigure}{.5\textwidth}
  \centering
  \caption{Shift over ONLY political context dimension}
  \includegraphics[width=\linewidth]{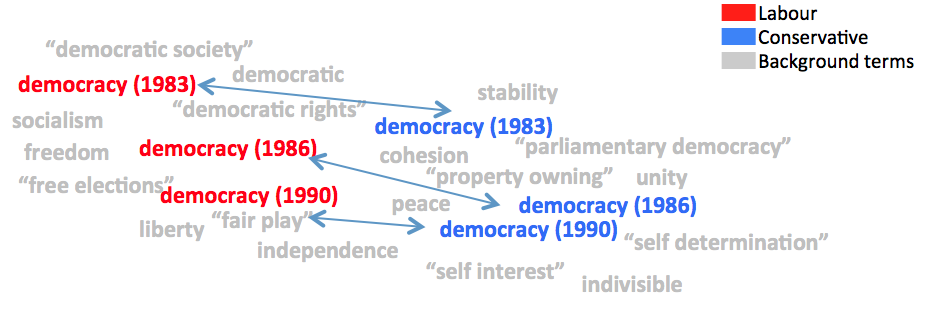}
  \label{fig:sub2}
\end{subfigure}
\vspace*{-\baselineskip}
\caption{Visualization of semantic shifts in meaning of words ``democracy" and ``moral" over time and along Conservative and Labour parties in the UK parliament. 
The approach proposed in \cite{Hamilton2016} is used for visualization. 
(a) The meaning given by Labours to ``moral'' is shifted from a ``philosophical'' concept to a ``liberal'' concept over time. In the same time, the meaning of this word is shifted from a ``spiritual" concept to a ``religious" concept from Conservatives' viewpoint. Moreover, two parties gave a very different meanings to this word. 
(b) The meaning of ``democracy" is stable over time for both parties. However, Conservatives refer to democracy mostly as a ``unity" concept, while Labours associate it with ``freedom" and ``social justice". 
}
\label{fig:intro}
\vspace*{-\baselineskip}
\end{figure*}

As an example, Figure \ref{fig:intro} shows the semantic shifts over two dimensions: time and political context, i.e. membership of a parliamentary party at the British House of Commons. The speeches given by the members of each party are used for constructing their corresponding semantic spaces. This can be extended to social parties or groups of like-minded people in social forums such as Facebook. The first example in the figure (the word ``moral") shows that a semantic shift can occur over time and across different contexts. However, as the second example shows, although the meaning of a word (such as ``democracy") can stay stable over time, it can still differ between certain groups. Therefore, social context is another valuable dimension that can explain  semantic shifts.
In this paper, we explore the \emph{semantic stability} of words by computing how contextual factors, such as social background and time, shapes---or, at least reflects---shifts in meaning.

We first use distributional semantics to generate embedding spaces from categorized corpora, where a category can be a certain context (such as speeches given by a political party). In the example given in Figure \ref{fig:intro}, there are two categories: Conservative and Labour parties. Then we propose different approaches to compare the vector representation of words between spaces. 
In the remainder of this paper, we define each of these categories as \emph{viewpoints}, since they reflect the semantic constellation of terms from a specific social perspective. 
In this paper we only consider two viewpoints. However, our approaches are easily extendable to multiple (i.e non-binary) viewpoints. 
The challenging part of this task, and the main contribution of this paper, is to develop techniques that compare vectors across spaces with different dimensionality structures. 
We consider three methods for comparing meaning across vector spaces.
(1) Inspired by \cite{mikolov-mapping-2013}, we create a linear mapping between two embedding spaces, project words from one embedding space to the other and measure whether the projected word lands closely to the word in the other space. (2) Inspired by \cite{Kenter2015}, for each viewpoint, we construct a graph such that the nodes are words and edges are the similarities between them. Then, using graph-based similarity measures we compute how similar the neighbors of a word in two embedding spaces are. (3) We define a measure that combines these two measures.

As stated, in this work our \emph{main research problem} is to study how semantic shifts in words are happening not just over time dimension but also social dimension, quantify the size of shifts, and explore the applications that  can benefit from the information about shifts. We evaluate the proposed approaches in three different tasks: measuring semantic shifts, document classification, and contrastive viewpoint summarization. 

Our main contributions are: (1) We show that semantic shifts not only occur over time, but also across different viewpoints in a short period of time. (2) We improve the linear mapping approach \cite{mikolov-mapping-2013} for detecting semantic shifts and propose a graph-based method to measure the size of semantic shifts in the meanings of words. (3) We employ word stability measures in contrastive viewpoint summarization and document classification and extensively evaluate our proposed approach  to these tasks. (4) Our analysis shows that the two laws of semantic changes proposed in \cite{Hamilton2016} hold for semantic shifts across viewpoints. Moreover, we introduce a new law of semantic changes which implies that concrete words are less likely to shift meaning while abstract words are more likely to do so. (5) We make the evaluation dataset for detecting semantic shifts and contrastive viewpoint summarization publicly available.\footnote{The datasets are available here: \url{http://dx.doi.org/10.7910/DVN/BJN7ZI}.} 

The remainder of this paper is organized as follows: after expanding on related research in \S\ref{relatedWork} we continue with describing our methods in \S\ref{sec:methods}. Then we explain the experimental setup and validation methods in \S\ref{experimentalSetup}. \S\ref{results} describes the merits and defects of each of the methods used, and proceeds with a detailed discussion of the validation. Finally, \S\ref{conclusion} concludes the paper with a brief discussion on the possible future directions.


\section{Related Work}
\label{relatedWork}
In this section, we review the related studies from three perspectives: detecting semantic shifts, methods for detecting ideology and approaches to viewpoint summarization.



\subsection{Detecting semantic shifts}
With the appearance of Word2Vec \cite{mikolov2013efficient} and GloVe \cite{Pennington2014}, unsupervised methods have become increasingly popular as tools for generating vector representation of words.
Notwithstanding the popularity of these vector representations, relatively few studies have attempted to compare embeddings generated from different corpora. 
%
The approaches closest to our approach are \cite{Jatowt:2014:FAS:2740769.2740809, Hamilton2016}. 
\citet{Jatowt:2014:FAS:2740769.2740809} create time-stamped word representations per decade, and use these to monitor semantic fluctuations over more than 400 years.
Words are represented as high dimensional vectors in which the values indicate how often a word co-occurs in the close vicinity of the target word.
\citet{Hamilton2016} use orthogonal Procrustes to align embeddings learned for different time-periods. They show that using a linear transformation is effective to find semantic shifts over time. Moreover, based on their proposed method for measuring semantic shifts over time, they propose two laws of semantic change.
Similar to these works, other studies also tried to capture semantic shifts in the meaning of words over time \cite{Kenter2015, Kulkarni2015, hamilton2016cultural, Ho2016, Yao2017} and also in the meaning of loanwords \citep{Takamura2017}.   

An alternative to our embedding based approach would be to use of a direct high-dimensional representation of the co-occurring terms as in e.g., \cite{Jatowt:2014:FAS:2740769.2740809, DBLP:conf/conll/LevyG14} which retains the dimensionality structure and allows a direct comparison across vector spaces. 
However, given that we want to detect semantic differences, it would be unrealistic to assume that all the dimensions mean the same in both corpora.

Another relevant line of study is monitoring and tracking events and topics over time \cite{Huang2017, Vaca2014, Li2012}. These approaches are aiming at detecting a set of topics and monitoring their change over time. Our approach is different than topic tracking methods as we do not restrict ourself to monitoring a limited set of topics. 
For evaluating the proposed approach, similar to previous work in detecting semantic shifts over time \citep{Kenter2015, Kulkarni2015, Hamilton2016, hamilton2016cultural, Jatowt:2014:FAS:2740769.2740809}, we select a small set of words whose meaning shifted and evaluate how the proposed approach is successful in detecting them. 

\subsection{Ideology and political text classification}
Besides monitoring changes in meaning, this paper demonstrates how knowledge about semantic shifts contributes to other tasks such as the classification of political texts. 
\citet{Kusner2015} applied word embeddings to calculate the distance between documents and utilized these estimated distances for classifying documents. 
Their results show that embedding-based approaches to document classification outperform others such as LDA and LSI.
Similar to previous work, we utilize word vectors for text classification.
Our task differs, however, since we employ multiple embeddings to enhance classification performance.

We use political text to evaluate the proposed approaches. 
Previous approaches to political text classification \cite{Hirst2014, Dehghani2016-ICTIR, Dehghani2016-CLEF, Dehghani2015-ECIR} are largely limited to word counting---or other units such as syntactic rules---thereby ignoring the adversarial semantics that characterize political discourse. 
Using word embeddings we attempt to capture this `macrocosmos' of political ideas. 

Our approach for document classification is to use word embeddings and expand documents using extracted associations that are specific for each class. 
This kind of document expansion helps in resolving the vocabulary mismatch issue and increasing the discrimination between different classes. Using word embeddings was shown to be very effective to boost the performance in document classification \cite{Kusner2015, Xing2014, Peng2016}.  

\subsection{Cross-perspective opinion mining \\ and summarization}
Even though opinions can be extracted at the word, sentence, or document level \cite{Pang2008}, they are usually represented as topics.
Most of the current approaches rely on topic models and jointly extract topic-opinion pairs from a set of unlabeled documents \citep{Lin2009, Mei2007, Ren2016}. 
They consider, however, only one point of view about topics. 
To extract contrastive opinions about topics, previous research \citep{Fang2012, Kim2009, Thonet2016} proposed to jointly extract topics and opinions coming from different viewpoints. 
Besides extracting polarity score of each viewpoint about topics, these approaches also summarize the opinions about topics. 
Similarly, we also perform \textit{contrastive viewpoint summarization} and for each topic we estimate a score which expresses and summarizes differences in word meaning. 
However, instead of opinions, we estimate and summarize the diverging viewpoints on a concept. Viewpoints are different than opinions as they do not necessarily carry sentiment information. 


\section{Measuring Word Stability}
\label{sec:methods}
In this section, we describe our approach for measuring semantic stability of words.
\subsection{Task overview}
We define semantic \emph{word stability} as the similarity of (a word's) vector representation across viewpoints. A \emph{viewpoint} is defined as a set of texts that share a specific metadata feature, for example texts generated by a social entity such as a political party.
Words whose meaning is independent of perspective will obtain a high stability score---for example in a political context we expect conservatives and progressives to disagree on the concept of `democracy', but not on the semantics of the word `lettuce'.
More formally, our method takes as input a word, and returns a number that expresses its meaning stability across viewpoints.

To measure the semantic stability of words, we first use distributional semantics to create a separate embedding space for each viewpoint. 
Then, using trained embeddings we map each word to a vector in their respective viewpoint. Finally, we compare the embeddings of each word in different embedding spaces. Our proposed approaches are applicable for words in the intersection of the vocabularies of two embeddings spaces.
In the reminder of this section, we use $V^0$ and $V^1$ to represent the created embedding spaces for two viewpoints. 
 $V^i_w$ is the vector of word $w$ in embedding space $V^i$.

Because the embedding spaces are different and have different dimensionality structures, we cannot compare the vectors of a word in two different spaces directly. In this paper, we propose different approaches to address this issue.
Below, we describe three methods for comparing words in different vector spaces: linear mapping, neighbor-based approach and, lastly a combination of the two.\\

\subsection{Linear mapping}
\label{sec:mapping}
The application of linear transformation for translating vectors from one space to another was first proposed by \cite{mikolov-mapping-2013, Hamilton2016}. 
In this approach a set of words with their extracted vectors in two embedding spaces are used to learn a mapping.
\citet{mikolov-mapping-2013} start with a set of training words (mainly function words, whose meaning should be stable irrespective of viewpoint or domain) in two embedding spaces. Using the training samples, the goal is to learn a transformation matrix $W^{ij}$ from embedding space $i$ to embedding space $j$ that minimizes the distance between the words and their mapped versions.
The transformation matrix is learned using gradient descent algorithm.  
The objective function is:
\begin{equation}
\underset{W^{ij}}{\mathrm{argmin}} \sum_{w \in X} \norm[\bigg]{W^{ij}  V^i_w -  V^j_w}^2,
\end{equation}
where $X$ is the set of training words. We denote the transformation matrix from embedding space $V^i$ to embedding space $V^j$ by $W^{ij}$.
We use a standard stopword list with a few additional words added (very frequent words) to learn the transformation matrix. As the meaning of these words should in theory be similar in both time periods, they serve as fixed points in the mapping around which the words with varying meaning are situated. The transformations are learned on a total of 813 words from the stop list. 

`Stability' measure of a word $w$, is then expressed by the following measure:\\
\begin{equation}
\label{eq:oneway}
s_{lin}(w) = \frac{sim_{01}(w) + sim_{10}(w)}{2},
\end{equation}
\begin{equation}
\label{eq:twoway}
sim_{ij}(w) = cos(W^{ji}W^{ij}V^i_w, V^i_w ),
\end{equation}
\noindent
where $cos$ is the cosine similarity. The stability of a word using this measure equals to the similarity of its vector to its mapped vector after applying the mapping back and forth.

\subsection{Neighbor-based approach}
\label{sec:neighbors}
The second method for measuring word stability is based on the intuition behind graph-based node similarity measures. The similarity of two nodes in a graph is determined by the similarity of their neighbors \cite{Jeh2002}. We consider each word in an embedding space as a node and its neighbors are the closest nodes to it, measured by cosine similarity. However, instead of one graph, we construct two graphs for two embeddings. For each word, we calculate the similarity of its neighbors in two different graphs and use this similarity as the stability of the word. 
This method assumes that words with similar meaning have similar neighbors. Thus, we can calculate stability by quantifying the extent to which words in different spaces still share neighbors. 

Based on this assumption, we define an iterative method for calculating word stability. The algorithm is described in Algorithm \ref{alg-neigh}.
We first suppose that all words are stable and initialize $s_{nei}^0(w)=1$ for all words. Secondly, depending on the depth parameter $t$, this method also takes into account the ``neighbor's neighbor'' etc. At first iteration, only direct neighbors contribute to the stability of words. At iteration $t=k$, the indirect neighbors accessible by $k$ edges in the graph contribute to the stability of words. 

\begin{algorithm}[t!]
\SetAlgoLined
\KwIn{$V^0$:  embedding space of viewpoint 0} 
\KwIn{$V^1$: embedding space of viewpoint 1}
\KwIn{$T$: the number of iterations} 
\KwIn{$\mathcal{V}$: the intersection of the vocabularies of $V^0$ and $V^1$}
 \KwResult{$s_{nei}^T$: a vector containing the stability of words}
 \For{w $\in$ $\mathcal{V}$}{
  $s_{nei}^0(w) = 1$
 }
 \For{t $\leftarrow$ 1 to $T$}{
 \For{w $\in$ $\mathcal{V}$}{
 $\mathit{sim}^t_{01}(w) = \frac{\sum_{w'\in N^1_w} cos(V^0_w, V^0_{w'}) s_{nei}^{t-1}(w')}{|N^1_w|}$\\
 $\mathit{sim}^t_{10}(w) = \frac{\sum_{w'\in N^0_w} cos(V^1_w, V^1_{w'}) s_{nei}^{t-1}(w')}{|N^0_w|}$\\
  $s_{nei}^t(w) = \frac{sim^t_{01}(w) + sim^t_{10}(w)}{2}$\\
  Min-Max normalize $s_{nei}^t$ to fall into [0,1] interval\
 }
}
 \caption{The algorithm for computing Neighbor-based stability of words. $N^0_w$ is the set of most similar words to $w$ in embedding space $V^0$ based on cosine similarity of words vectors.}
 \label{alg-neigh}
\end{algorithm}

\subsection{Combination: Co-occurrence of neighbors \\ and linear mapping}
\label{sec:comb}
The third, and last, stability-metric combines the neighbor-based approach with linear mapping. 
Each of these metrics are providing different signals about the stability of a word: linear mapping is solely based on the mapped vectors of the word while the neighbor-based approach is based on the vectors of neighbors and does not take into account the vector of the word itself. Thus, we combine these metrics to achieve better stability scores.
This stability measure is based on the number of co-occurring neighbors \textit{and} their similarity to the target word. The algorithm is described in Algorithm \ref{alg-com}. 
For each word $w$, the weights of its neighbors reflect their place (or index) in a ranked list comprising the $N$ most similar words to $w$. We define and combine two different stability signals: 1) $C^t_{ij}(w)$ represents the count of neighbors of word $w$ in embedding $V^i$ based on their index in the ranked list of neighbors of $w$ in embedding $V^j$.  $C^t_{ij}(w)$ is defined based on the words which are neighbors of word $w$ in both embedding spaces. 2) $sim_{ij}^t(w)$ is based on similarity of mapped vectors from embedding space $V^i$ to embedding space $V^j$ and their vectors in space $V^j$, for the words that are neighbors of $w$ in embeddings $V^i$ but not in embedding $V^j$.

To give an example of how to compute $C_{ij}^t(w)$ in Algorithm \ref{alg-com}, consider the following neighbor list:
\begin{align*}
N^0_w = [n1,n2,n3,n4,n5]\\
N^1_w = [n2,n4,n1,n5,n6]
\end{align*}
Each neighbor in list $N^0_w$ is obtained (if possible) from list $N^1_w$, along with the index. The final count after the first iteration ($C_{01}^0(w)$) then becomes: $C_{01}^0(w) = 5*4 - (2+0+1+3) = 14$.
Note that this summation contains four terms instead of five, as neighbor $n3$ does not occur in list $N^1_w$. Therefore, in order to be able to take neighbor $n3$ into account when computing the agreement, the linear mapping is used to map the vector of $n3$ to a vector representing it in $V^1$. Then the cosine similarity from the mapped vector to the target word vector is incorporated in calculating the stability value of $w$ (using $sim_{01}^0(w)$). 
$\lambda$ is defined as follows: 

 \begin{equation}
 \label{eq-comb}
 \lambda = \begin{cases}1 \textnormal{,  \qquad   $N^0_w = N^1_w$} \\
 0 \textnormal{,  \qquad  $C_{01}^t = 0$ and $C_{10}^t = 0$} \\
 0.5, \enspace \quad otherwise\end{cases}
 \end{equation}
 
 \begin{algorithm}[t]
\SetAlgoLined
\KwIn{$V^0$:  embedding space of viewpoint 0} 
\KwIn{$V^1$: embedding space of viewpoint 1}
\KwIn{$T$: the number of iterations} 
\KwIn{$\mathcal{V}$: the intersection of the vocabularies of $V^0$ and $V^1$}
\KwIn{$\lambda$: the combination parameter determined by Equation \ref{eq-comb}} 
 \KwResult{$s_{com}^T$: a vector containing the stability of words}
 \For{w $\in$ $\mathcal{V}$}{
  $s_{com}^0(w) = 1$
 }
 \For{t $\leftarrow$ 1 to $T$}{
 \For{w $\in$ $\mathcal{V}$}{
  \For{i, j $\leftarrow \{0, 1\}$ $\wedge i \neq j$}{
  $C_{ij}^t(w) = |N^i_w|\times |N^i_w\cap N^j_w| - \sum_{w' \in N^i_w\cap N^j_w}\  \frac{rank_j(w')}{s_{com}^{t-1}(w')}$\\
  \vspace{5mm}
  $sim_{ij}^t(w) = \frac{\sum_{w' \in \{N^i_w \setminus N^j_w\}} cos(W^{ij}V^i_{w'}, V^j_{w}) s_{com}^{t-1}(w')}{|N^i_w\setminus N^j_w|}$\\
  }
  $s_{nei}(w) = \frac{C_{01}^t(w)+C_{10}^t(w)}{2\sum_{i=1}^{N_w}i}$\\
  $s_{lin}(w) = \frac{sim_{01}^t(w) + sim_{10}^t(w)}{2}$\\
  $s_{com}^t(w) = \lambda s_{nei}(w) + (1-\lambda) s_{lin}(w)$\\
  Min-Max normalize $s_{com}^t$ to fall into [0,1] interval\\
 }
}
 \caption{The algorithm for computing the stability of words based on combination of neighbor-based and linear mapping approaches. $|N_w|$ is the number of neighbors considered (i.e. 100), and $rank_j(w')$ is the rank that neighbor $w'$--- which is an element of $N_i(w)\cap N_j(w)$--- has in the ranked list of neighbors of $w$ in embedding space $V^j$.}
 \label{alg-com}
\end{algorithm}


\section{Experimental Setup}
\label{experimentalSetup}
We evaluate the performance of our approach intrinsically detecting semantic shifts task (the details of this evaluation method is described in \S\ref{intrinsic} and \S\ref{RQ1.1Res}) and extrinsically in document classification and viewpoint summarization tasks (the details of these evaluation methods are described in\S\ref{extrinsic} and \S\ref{classificationRes}). Our main research questions are:
\begin{description}
  \item[RQ1]
    How effective are the proposed approaches in quantifying the changes in word meaning over various dimensions such as time and political context?
    \item[RQ2] 
    To what extent can these models improve performance on other tasks, such as  document classification?
   \item[RQ3]
    How do the proposed approaches perform in summarizing different viewpoints expressed in two large corpora about different concepts?
   \item[RQ4]
   Do temporal laws of semantic change hold for shifts across viewpoints? 
\end{description}

RQ1 is concerned with the quality of stability values estimated for words using different approaches. To answer RQ1, we construct an evaluation set and evaluate the accuracy of different approaches in measuring stability of words. 
In \S\ref{RQ1.1Res} the results of the experiments regarding RQ1 are reported.

To answer RQ2, we use the stability values for document classification. We first expand the documents using the stability values and employ the expanded documents for classifying the speeches in the UK parliament to the parties. The details of this experiment are described in \S\ref{documentExpansion} and the results are reported in \S\ref{classificationRes}. 

To answer RQ3, we utilize the word stability values for contrastive viewpoint summarization. We first generate the summary for a set of chosen words using different methods and ask human annotators to assess the summaries. The details of the evaluation process are described in \S\ref{summarizarionGroundTruth}. The results of experiments related to RQ3 are reported in \S\ref{summarizationRes}

RQ4 is concerned with the validity of laws of semantic shifts across viewpoint. To answer RQ4, we analyze the correlation of semantic shifts with their frequency, polysemy, and concreteness. The results of the experiments concerning RQ4 are described in \S\ref{RQ4}.

\subsection{Datasets}
\label{datasets}
To evaluate how effectively the methods described in \S\ref{sec:methods} capture and summarize semantic shifts, we run multiple experiments using data sourced from the New York Times corpus\footnote{https://catalog.ldc.upenn.edu/LDC2008T19} and the digitized proceedings of the British House of Commons--- also referred to as the Hansard\footnote{ http://www.parliament.uk/business/publications/hansard/commons/}.

Our corpus of political texts comprises the parliamentary and public speeches from the Thatcher years.
 This period contains 640,184 speeches. 
Within the broader context of British postwar politics, this era represents a break with the postwar Keynesian consensus, and was accompanied by a hardening division between left and right.  
In this paper we study how much the concepts in the Thatcher period have different meaning from a `Conservative' and `Labour' point of view. 

The New York Times dataset contains 1,855,671 articles published between 1987 and 2007. We study how the meanings of words shifted after 9/11 in this newspaper. For example, as the terrorists involved in the 9/11 attacks were professors of Islam, it could be of value to investigate whether this had any affect on how Islamic faith is framed in media discourse. To do so, we divide the articles in the New York Times dataset into two viewpoints, i.e. articles before and after 9/11. We consider these two sets as two different viewpoints and study how the meaning of concepts are different based on these two point of views. 

\subsection{Preprocessing and general setting}
We use Word2Vec \cite{mikolov2013efficient} to generate word embeddings. We apply Skipgram architecture and remove words with less than 20 occurrences. We train an embedding with 300 dimension with a window size of 10.

\textbf{Linear mapping} refers to the linear transformation method introduced in \S\ref{sec:mapping}. \textbf{Neighbor-based} method is the method introduced in \S\ref{sec:neighbors} and \textbf{Combination} is the method described in \S\ref{sec:comb}.
In estimating stability values using the Combination methods, we set $|N_w^i|=100$ which reflects that we only use top 100 closest word to each word for estimating the stability values. In Algorithm \ref{alg-neigh}, for each word $w$, we again use top 100 closest word to each word for estimating the stability values, however from this set we remove neighbors with similarity lower than 0.4 to $w$.
For calculating stability values we set $T=5$ (the number of iterations of the Neighbor-based method and the Combination method) since based on our experiments after 5 iterations the stability values do not change considerably. 
The linear mappings are created using Gradient Descent algorithm with a maximum number of 50,000 iterations and a learning rate of 0.01.
Before creating embedding spaces, we use the method proposed in \cite{Mikolov2013} to detect bigrams. We consider documents as a combination of unigram and bigrams terms.

\subsection{Intrinsic evaluation}
\label{intrinsic}
In this section, we describe the dataset we use for evaluating our approaches in detecting semantic shifts.

\textbf{Ground truth for semantic shifts}
\label{wrodStabilityGroundTruth}
Following previous work \citep{Kenter2015, Kulkarni2015, Hamilton2016, hamilton2016cultural, Jatowt:2014:FAS:2740769.2740809}, we create a small  dataset to evaluate the performance of the proposed approaches. Because we do not possess text-book definitions to evaluate our model--the meaning of the words we study are contested by politicians and academics alike--we assess whether the representations we extract tie in with the perceptions of experts.
 To validate our method, and see how well we do in the replicating diverging interpretations on political concepts, we choose 24 words which we know were central to many of the controversies of the Thatcher era (1979-1990) and ask experts whether they could recognize the viewpoint. The selected concepts are shown in Table \ref{table:groundTruth}. 
\begin{table*}[t]
	\centering 
	\caption{\label{table:groundTruth}
          The selected concepts for evaluating the word stability measures in detecting semantic shifts and summarizing viewpoints.} 
          \vspace*{-\baselineskip}
	\begin{tabular}{c c}
    \toprule
    \textbf{Detecting semantic shifts task (selected from the UK parliament)} & \textbf{Summarization task (selected from the New York Times)} \\
    \hline
    privatisation, unemployment, working\_class, society& islam, muslim, fundamentalism, radicalized \\
    homosexual, fairness, public\_sector, justice, liberalism & wtc, terrorism, terrorist, terrorist\_attacks \\
    communism, constitution, free\_market, sovereignty & ground\_zero, hijacking, terrorist\_targets, security\\
    accountability, inequality, moral, conservatism, profit & anti\_terrorism, anti\_americanism, 911, airport\\
    morality, tolerance, opponent, poor, bureaucracy, rich &\\
\bottomrule
	\end{tabular}
	\vspace*{-\baselineskip}
\end{table*}

The words we select for evaluation, reflect the prevalent debates of Thatcher period described in \S\ref{datasets}, and focus on issues such as economic reform, labour disputes and equality. For each word we select the most Conservative and Labour associated terms, thus discarding the overlapping or shared items.  These lists are obtained from embeddings trained on a corpus containing speeches from either Conservative or Labour members. For each word, we select its most similar neighbors in the two embedding spaces and create two lists. These two lists of related terms are then anonymized--meaning that we remove the party where the list stems from-- and given to experts, whom we asked if, when shown a concept like `democracy', they could identify which list described the Conservative or Labour interpretation. 
There were 4 annotators who were all political scientists and familiar with the political history of UK. None of the authors participated in the annotation. All annotators annotated all 24 words. The agreement between the  annotators, based on Fleiss' Kappa, is 0.47 ($p-value < 0.001$) and the overall accuracy is 0.75, indicating that they were able to detect the correct labels in most of the cases.
Upon closer inspection, the low agreement may result from the fact that the summaries send mixed signals. 
The concept `homosexuality', which was mislabelled by all respondents, is a good example. While the Labour party, at the end of the eighties, was largely supportive of gay rights, the Conservatives took a more negative stance, which led to the infamous Section 28 of the Local Government Act (1988). The phrase `promoting homosexuality' was as a Labour feature, and could be interpreted as reflecting a more positive opinion, but the same words also figured in the conservative Act, albeit prefixed with `not'. 
In general, the summaries fail to capture whether the associated words are in a synonymic or antonymic relation with the target concept, which significantly complicated the interpretation.

\subsection{Extrinsic evaluation}
\label{extrinsic}
In this section, we describe the datasets and approaches used for evaluating the proposed word stability measures in document classification and contrastive viewpoint summarization tasks. 

\subsubsection{Document classification: methods and metrics}
\label{documentExpansion}
We evaluate our stability measures by employing them in the task of ideology detection. The input in this task is a speech held in the UK parliament and the task is to determine the party (the ideology) of the speaker. 
We train an SVM classifier on a collection of speeches, categorized as either Labour or Conservative. 
We aim to optimize classifier accuracy by expanding documents as follows: we want to amplify the fact that a speech belongs to a certain class by adding for each unstable word in the speech its top $n$ most similar but unstable words in the embedding space belonging to this class. This is reminiscent of the idea behind doc2vec \cite{le2014}, with the difference that we explicitly change the document. Note that we only expand the documents in the training set, not in the test set.  
This setup has two parameters. The first is the threshold $\theta$ which categorizes all terms as either `stable' or `unstable' depending on their stability value. We will optimize this parameter in our experiments using a development set. The second parameter is the number of terms to add for each unstable word. As usual in expansion setups adding too few and too much will lead to worse performance. The effect of $n$ on the performance of different classifiers is shown in Figure \ref{figure-param}.

We discard speeches of less than 50 words and then randomly select 50,000 speeches from the Thatcher period for each of the parties for performing classification. The mean and median length of the selected speeches are 282 and 107 words, respectively.

We do 10-fold cross validation and report Precision, Recall, and $F_1$ measures in the classification task. We use 8 folds as training data, one fold as development set to tune $\theta$, and one fold for testing. For each document we construct a feature vector using TF-IDF values. Each element of this vector corresponds to a word and its value is TF-IDF weight of the word in the document normalized by the length of the document. After expanding documents in the training set, we re-compute TF-IDF values for words. 
 
 As a baseline, we compare the performance against a different expansion method, which inserts neighboring words calculated from an embedding trained on the whole corpus excluding test documents. 
 Since we do not have stability values for this method, we expand all words in speeches using the general embedding.
 We refer to this method as `SVM+General' in \S\ref{results}. The general word embedding is trained on all speeches from the Thatcher period. We compared `raw SVM' to `SVM+General' and the latter performed better. Therefore, we take `SVM+General' as our baseline.

\subsubsection{Contrastive Viewpoint summarization: dataset, methods, and metrics}
\label{summarizationMethod}
In this section we describe the evaluation set and our approach for evaluating word stability measures in the contrastive viewpoint summarization task.

\textbf{Contrastive viewpoint summarization} 
We use the estimated stability values to summarize viewpoints about concepts. The input is a concept $w$, the stability values estimated using the approaches introduced in this section and the length of the summary $l$. The output is two lists of summaries in which each list contains $l$ words describing a viewpoint about $w$.
 
To summarize a viewpoint $V^i$ about a given concept $w$, we take the top 100 most similar words to $w$ in embedding space $V^i$. Then starting from the most similar neighbor, the word is added to the summary if a neighbor is in the overlapping vocabulary of the two embedding spaces and if the stability of the neighbor is equal to or below the set threshold. This process is continued and $l$ words are selected as the summary. As the top 100 neighbors are ordered from highest similarity to lowest, the summaries will follow the same trend. In this task, we set the length of the summary to 5 and the number of iterations of the Neighbor-based method and the Combination method to one (i.e. we only use direct neighbors).

\textbf{Ground truth for viewpoint summarizarion }
\label{summarizarionGroundTruth}
The summaries produced using the three summarization methods are assessed through peer evaluation. To make the evaluation set, we use the New York Times dataset. We study which method best summarizes the shift in meaning after 9/11.  
We select a total of 16 concepts, which are chosen based on relevant literature.
The selected concepts are shown in Table \ref{table:groundTruth}.
Summarisation questions consist of a concept and its accompanying summaries before and after 9/11 for all three summarisation methods.
For each concept, the question is as follows: 
\emph{In your opinion, which of the summaries belong to the given concept `Before 9/11' and which of the summaries belong to `After 9/11'?} 
No specifications regarding how many summarizations per category were given per concept, leading to a fairly open evaluation.
Questions were randomized per survey, as were the options for the summarisation questions. Each summary was annotated by 10 people. The agreement between the  annotators, based on Fleiss' Kappa, is 0.54 ($p-value < 0.001$).
Before asking annotators, we have the labels of the summaries (before 9/11 and after 9/11). A good summary is the one that annotated by 10 people correctly. Therefore, the number of times the label of a summary generated by a particular method is annotated correctly shows the performance of the method in summarization task. In \S\ref{results} we report the performance of different methods as the number of times the annotators detected the labels correctly in terms of Precision, Recall, and $F_1$ of the annotators on the generated summaries. 

We use the New York Times dataset instead of the parliamentary proceedings from the Thatcher period for the summarization task. The reason is that it is more straightforward for our annotators to assess whether the summary of a viewpoint about a word belongs to before or after 9/11 event, compared to assessing whether the summary belongs to the Conservative or Labour party.

\subsection{Statistical significance}
For statistical significance testing, we compare our methods to baselines using paired two-tailed t-tests with Holm-Bonferroni correction for multiple hypothesis testing. We set $\alpha$ (the desired significance) to $0.05$. 
In \S\ref{results}, $^\blacktriangle$ and $^\blacktriangledown$ indicate that the corresponding method performs significantly better and worse than the corresponding baseline, respectively.


\section{Results}
\label{results}
In this section, following four research questions described in \S\ref{experimentalSetup}, we report the results of different word stability measures. 

\subsection{Results of word stability measures in \\ detecting semantic shifts}
\label{RQ1.1Res}
\begin{figure}[t]
\centering
\includegraphics[width=\columnwidth]{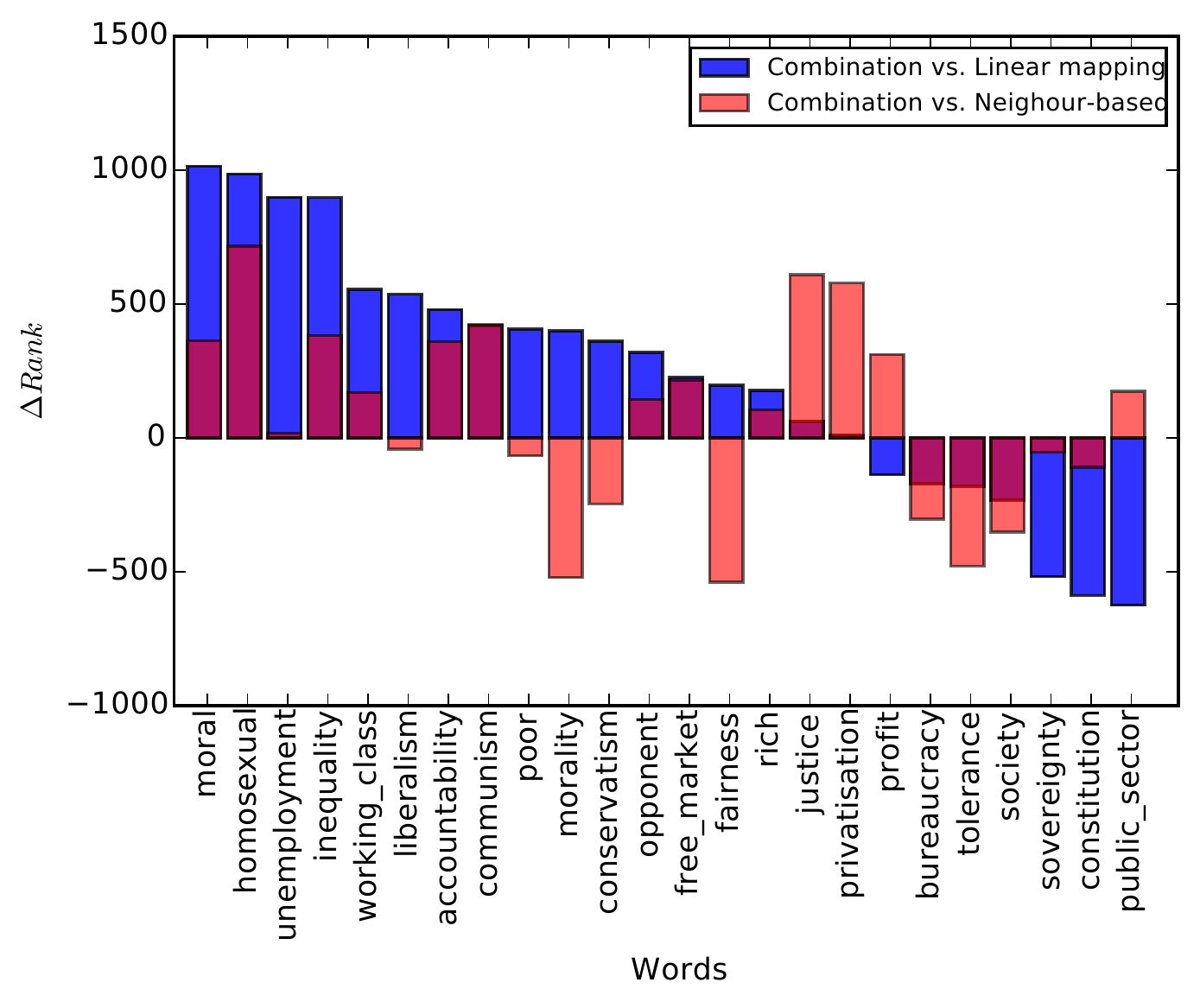}
\vspace*{-2.5\baselineskip}
\caption{The delta between the rank of the selected words in the rankings created by the linear mapping method and the Combination method and the rankings created by the Neighbor-based method and the Combination method.}
\label{figure1}
\end{figure}

To answer \textbf{RQ1}, we use the dataset described in \S\ref{wrodStabilityGroundTruth}. This dataset contains 24 words which are expected to exhibit ideologically divergence. The setup of this experiment can be found in \S\ref{intrinsic}.
We rank all words in the vocabulary based on the reverse of their stability values (unstable words are ranked higher). A good stability measure should rank the selected words higher.
 The average rank of the selected words in the ranking created using the Combination method is 462. Based on a paired two-tail t-test, this value is significantly lower than the one for the linear mapping method which is 681.  This shows that the proposed approach is effective in finding the words which have different meanings in the UK parliament.
  Figure \ref{figure1} shows the delta between the rank of the selected words in the rankings created by the two other methods and the Combination method. As can be seen most of the words are ranked higher by the Combination method compared to the other methods.

\begin{figure}[h]
\centering
\includegraphics[width=0.9\columnwidth, height=5cm]{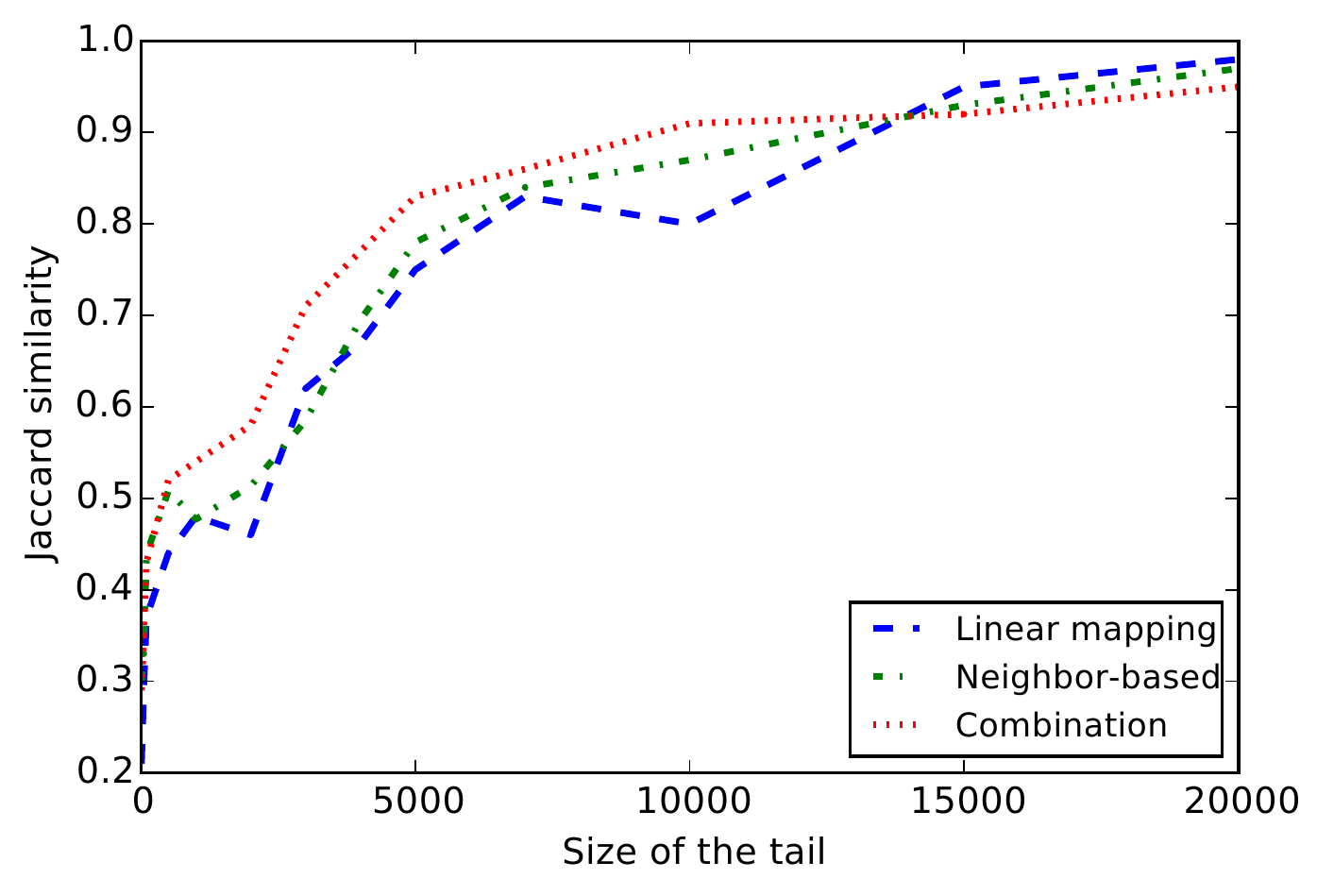}
\vspace*{-1.5\baselineskip}
\caption{Jaccard similarity of tails of the rankings created for the Thatcher and the Blair period using linear mapping, Neighbor-based, and Combination methods.}
\vspace*{-0.5\baselineskip}
\label{figure-tails}
\end{figure}

We run an additional analysis to see if our methods are robust with respect to semantically stable words. Specifically we assess if our approaches can detect words that do not change when moving from one party to another.  
For comparison, we also compute scores using speeches from the Blair period (1997-2007) and compare the tail of the ranking with the tail of the ranking created on the speeches in the Thatcher period. The main intuition is that if a word is stable, its meaning should not change over time (across different periods of the parliament). 
Figure \ref{figure-tails} shows the Jaccard similarity of the tails of the two rankings for various tail sizes across all methods. By increasing the size of the tail, more words are included and the intersection of the two lists and the Jaccard similarity are increasing. As can be seen, the Combination method has higher Jaccard similarity values in most of the cases. 
The Jaccard similarity of the tails when we set the tail size to 5000 (the size of the intersection of `Labour' and `Conservative' vocabularies is about 50,000) for the Combination method is 0.83 which is a high value. This value is 0.78 for the Neighbor-based approach and 0.75 for the linear mapping.

Table \ref{table:example} shows the head and the tail of the rankings of words based on instability values estimated for each of the used approaches. As can be seen, all approaches are good in finding highly stable words (the tails), as the tails of the ranking contain very general words which are expected to show little variation across viewpoints. However, the head of the list created by the linear mapping approach contains mostly words that we did not expect to shift such as `north' and `oil'. Unstable words in the Neighbor-based method's list such as `socialist' and `democratic' are expected to vary. This method is effective in finding these words. However, there are words such as `noble' and `space' in top of this list. 
Based on our analysis, the Conservatives included more aristocratic members (which are addressed as `noble' Friend) while Labour MPs use `noble' as a more quality.
Also, Conservatives use the word `space' when they refer to `space technology'. However, Labour use the word `space' to mostly speak about `living space or urban space'. Therefore, these two words do diverge and two parties use these words in different contexts to describe different concepts, but the relationship with ideology is not always straightforward. 

\begin{table}[t]
	\centering 
	\caption{\label{table:example}
          The head and the tail of ranking of words achieved using different word stability measures. For the Neighbor-based and the Combination methods the number of iterations is set to 5.} 
          \vspace*{-\baselineskip}
	\begin{tabular}{l c c}
    \toprule
    Method & Head & Tail \\
    \hline
     \multirow{5}{*}{Linear mapping} & gas & member\\
     & nuclear\_power & tuesday \\
     & north & thursday \\
     & oil & thank \\
     & church & nothing\\
     \hline
     \multirow{5}{*}{Neighbor-based} & noble & wednesday\\
     & socialist & friday \\
     & illegal & monday \\
     & democratic & tuesday \\
     & space & december\\
     \hline
     \multirow{5}{*}{Combination} & legislative & about\\
     & inequality & tuesday \\
     & private\_enterprise & side \\
     & noble& nothing \\
     & democratic & thursday\\
    \bottomrule
	\end{tabular}
\end{table}

From the results presented here we conclude that the Combination method is highly effective in detecting semantic shifts (as shown in Figure \ref{figure1}) and very robust with respect to semantically stable words (as shown in Figure \ref{figure1} and Table \ref{table:example}).

\subsection{Results of word stability measures in \\  document classification}
\label{classificationRes}
To answer \textbf{RQ2}, we use the method described in \S\ref{documentExpansion} for expanding speeches in the UK parliament during the Thatcher period and employ the expanded documents for classifying speeches by party. The setup of this experiment can be found in \S\ref{documentExpansion}.

Table \ref{table:1} shows the results of this experiment. In general, the results indicate that the proposed word stability measures help in discriminating documents.
Moreover, two other observations can be made from the results.
First, expanding documents, even with a general embedding can improve performance of the classifiers. 
Second, the Combination method performs better than the other approaches. The linear mapping approach does not outperform the baseline. The higher accuracy of the Combination method shows that, although the linear mapping approach does not improve the performance of the classifier, when it is combined with the Neighbor-based method, the performance is improved.

\begin{table}[t]
	\centering 
	\caption{\label{table:1}
          Results of classification of speeches to parties using different word stability measures. We consider SVM+General as our baseline.} 
          \vspace*{-\baselineskip}
	\begin{tabular}{l l l l}
    \toprule
    Method & \multicolumn{1}{c}{Precision} & Recall & \multicolumn{1}{c}{$F_1$} \\
    \hline
    SVM  & 0.781~~ & 0.718~~ & 0.758~~\\
    SVM+General &  0.793~~ & 0.735~~& 0.763~~ \\
    \hline
    SVM+Linear mapping & 0.804$^\blacktriangle$ & 0.738~~ & 0.770~~\\
    SVM+Neighbor-based & 0.823$^\blacktriangle$ & 0.768$^\blacktriangle$ & 0.795$^\blacktriangle$ \\
    SVM+Combination & 0.839$^\blacktriangle$ & 0.775$^\blacktriangle$ & 0.806$^\blacktriangle$ \\
    
    \bottomrule
	\end{tabular}
\end{table}

To gain additional insights about our approaches, we further analyze speeches which are correctly classified by the Combination method but not by the ``SVM+General".
The following (part of a) speech is an example of such samples:
\begin{small}
\begin{displayquote}
``...subsidise the residents of wasteful \textit{labour} authorities. If we were to strip away the surcharges and handouts, we would find that the \textit{labour} party's arithmetical inexactitude is almost a case for reference to the advertising standards authority. Having done that, we find that totally conservative areas have an average community charge of 305 pound, compared with the rip-off in totally \textit{labour} areas of 412 pound. Opposition members may think that this is a laughing matter, but a differential of no less than 107 pound per head for the privilege of voting \textit{labour} has a devilish impact on the charge payers of those areas.''
\end{displayquote}
\end{small}
This speech is given by a member of Conservative party. However, it is mostly about Labour party since mentions the `Labour' party occur more than 4 times. `Labour' is a very discriminative word for Labour party and that is the main reason that this speech is classified in Labour class. However, when we expand documents in the training step with words the Conservatives characteristically use to describe Labour members, the different sense of the word becomes apparent.
Thus, in the example mentioned above, the words such as `subsidize', `wasteful' and `inexactitude' will help more to classifying this example correctly. 

\begin{figure}[t]
\centering
\includegraphics[width=0.9\columnwidth]{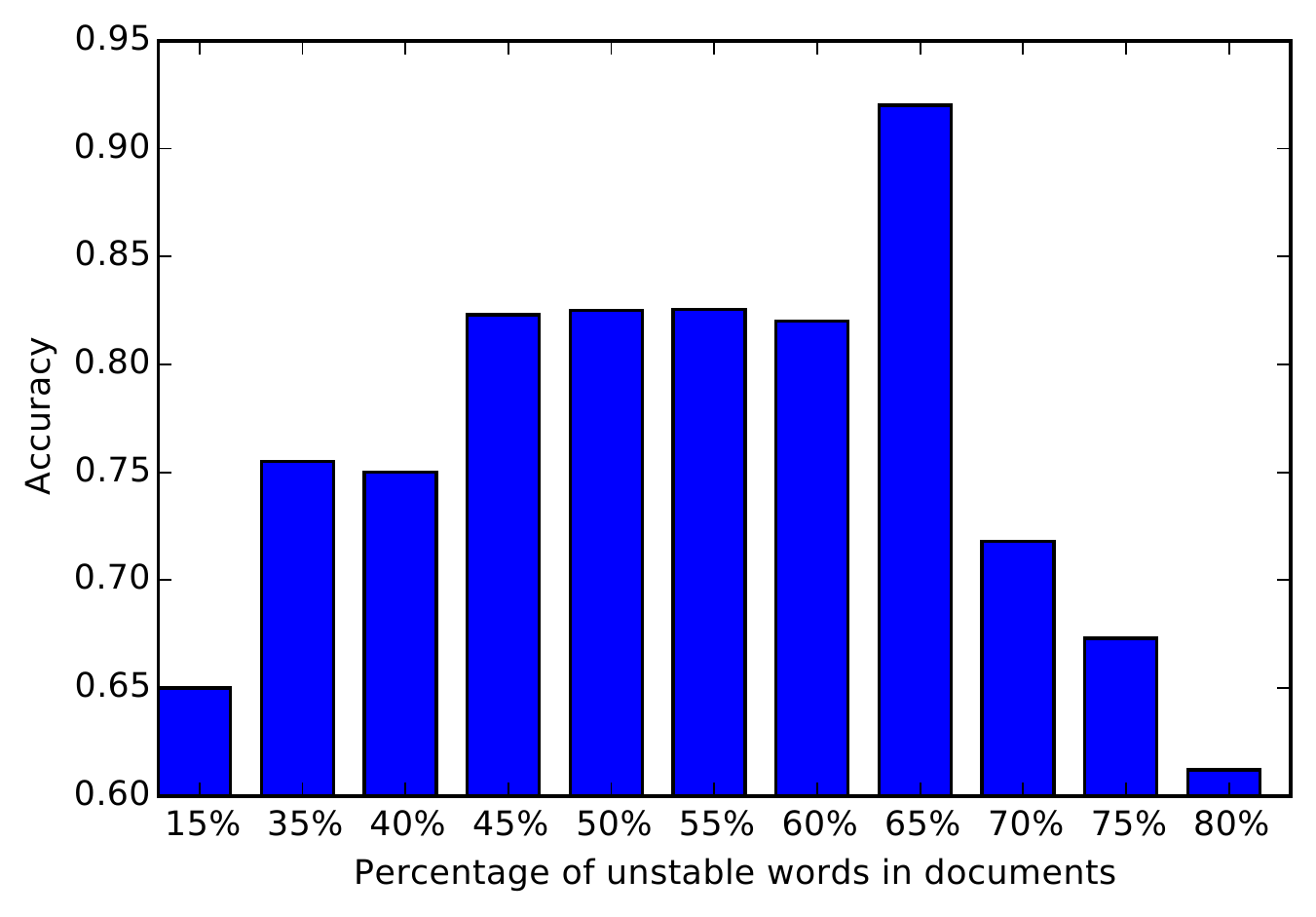}
\vspace*{-\baselineskip}
\caption{The accuracy of the Combination method in classifying the speeches in the Thatcher period for different levels of expansion. $\%i$ in x-axis is representing the documents that $\%i$ of their words are unstable.}
\label{fig:class-bins}
\end{figure}

Moreover, we analyze the speeches which are classified correctly by `SVM+General' and incorrectly using the stability measure. Our analysis show that most of these speeches are very short ones which do not contain any information about the author's viewpoint. When we filter out documents with less than 200 words length, the $F_1$ score of `SVM+General' is increased to 0.79 and the $F_1$ score of the Combination method is 0.85 and the improvement of the Combination method is more than the improvement of `SVM+General'. Another source of error is inaccurate stability values for words. This causes the expansion of documents with wrong words and lowers the accuracy of the classifier. Figure \ref{fig:class-bins} shows the accuracy achieved for different levels of stability values. We first calculate the percentage of unstable words (words for which their stability calculated using the Combination method is less than $\theta$) in documents. Then, we put the documents into different bins based on their percentage of unstable words and calculate the accuracy of the classifier for each bin. We only show the bins containing more than 1000 documents. For the highly unstable documents the accuracy is the lowest. This is mostly due to extreme expansion of these documents (since their words have low stability values) which are not accurate enough in most of the cases. The accuracy is higher when the stability value does not skew towards one of the extremes.

\subsection{Results of word stability approaches \\ in contrastive viewpoint summarization}
\label{summarizationRes}
This section answers \textbf{RQ3}. We use different word stability measures and the method described in \S\ref{summarizationMethod} to generate summaries for words. The setup of this experiment can be found in \S\ref{summarizationMethod}. To evaluate the performance of our methods in summarizing the viewpoints, we use the dataset described in \S\ref{summarizarionGroundTruth} and report results in Table \ref{table:2}.
In general, the performance of the Combination method is slightly better than the linear mapping approach. However, the difference is not statistically significant. The $F_1$ score achieved using the Combination method is 0.75 which is reasonably good and indicates that the annotators were able to detect the viewpoints using the provided summaries. 
The results show that the linear mapping method performs better than the Neighbor-based approach in the summarization task. This result is in contrast with the results achieved in the classification task. 
The summarization task is done on the New York Times dataset, while the classification task is performed on the UK parliamentary proceedings. In the parliamentary proceedings, the viewpoints are more apparent as neighbors of a given word generally serve as reliable descriptors of the viewpoint. Therefore, the Neighbor-based approach which is solely based on the similarity of the neighbors in two spaces performs better than the linear mapping method in the classification task. 

\begin{table}[t]
	\centering 
	\caption{\label{table:2}
          Results of different word stability measures in summarizing the viewpoints.} 
          \vspace*{-\baselineskip}
	\begin{tabular}{l c c c}
    \toprule
    Method & Precision & Recall & $F_1$ \\
    \hline

    Linear mapping & 0.74 & 0.74 & 0.74\\
    Neighbor-based & 0.66 & 0.64 & 0.65\\
    Combination & 0.75 & 0.74 & 0.75\\
    
    \bottomrule
	\end{tabular}
\end{table}

\subsection{Statistical laws of semantic change}
\label{RQ4}
In this section,we answer \textbf{RQ4}. Recently, \citet{Hamilton2016} proposed two quantitative laws of semantic change: 1) the law of Conformity which implies that ``the rate of semantic change scales with an inverse power-law of word frequency''. 2) the law of Innovation which reflects that ``the semantic change rate of words is highly correlated with their polysemy''.
To test if, besides accounting for change over time, these laws also help explaining semantic shifts across ideological perspectives, we used the UK parliamentary proceedings from the Thatcher period.
To check the first law, we construct two vectors (each entry in these vectors corresponds to a word and the length of the vectors are equal to the size of the intersection of vocabularies of Labour and Conservative parties): one using the frequency of words and one using their instability ($1-stability$). Then, we calculate the Pearson correlation between these vectors. 
To check the second law, again we construct two vectors in a similar way: one using the polysemy of words (we use WordNet to calculate the number of senses of words to quantify their polysemy) and one using the  instability of words. 

The Pearson correlation values are shown in Table \ref{laws}. The results show that: first, the law of conformity strongly holds using all measures. This becomes even more apparent when we use the linear mapping method. This is expected since we use highly frequent words for training the mappings. Second, the law of innovation does not strongly hold using all measures. We hypothesize that this is because even when parties use a word with low polysemy, they inject it with diverging meanings for example by using different sentimental words to express their opinion about the word. 

Moreover, we hypothesize that lexically different word senses are unlikely to appear in a short period, or in (still very similar) the political data we use. Thus, there are likely other, deeper causes such as concreteness of words. We study how the semantic change rate is correlated with the concreteness of words. Again, we construct two vectors: one using the concreteness of words and one using their instability. We use a dataset \cite{Brysbaert2014} containing the concreteness rating of words for constructing the concreteness vector.  The results are shown in Table \ref{laws}. The result indicate that there is a negative correlation between concreteness and instability and concrete words are less likely to shift. In fact, more abstract words are more likely to shift. 

\begin{table}[h]
    \centering 
    \caption{\label{laws} The Pearson correlation between the instability of words with their frequency, polysemy, and concreteness.} 
    \vspace*{-\baselineskip}
        \resizebox{0.45\textwidth}{!}{
        \begin{tabular}{l c c c}
        
    \toprule
     Measure & Conformity & Innovation & Concreteness\\
    \hline
    Linear mapping & -0.63 & 0.11 & -0.31\\
    Neighbor-based & -0.42 & 0.18 & -0.34\\
    Combination & -0.51 & 0.22 & -0.39\\ 
    \bottomrule
    
    \end{tabular}
    }
\end{table}

\subsection{Analysis}
In this section, we analyze the quality of the trained mappings between the embedding spaces and measure the effect of the word-expansion on the classifiers accuracy. 

\subsubsection{Quality of linear mappings}
To measure the quality of the created mappings, we report the average value estimated using Equation \ref{eq:twoway} over all words in the vocabulary. 
Table \ref{table:3} shows the results of this experiment. The average similarity calculated using one-way mapping ( $sim_{ij}(w) = cos(W^{ij}V^i_w, V^j_w)$) is low, meaning that when words are mapped from one space to the other, they are not close to the same word in the destination space. 
However, when we use Equation \ref{eq:twoway}, the average similarity value is high indicating that when going back to the same space, the word is mapped to its original vector. This shows that low value of similarity for one-way mapping is mainly due to instability of words which their meaning (and location in two spaces) are different based on different viewpoints.

\begin{table}[h]
    \centering 
    \caption{\label{table:3} Average cosine similarities of words after linear mappings on the UK parliamentary proceedings in the Thatcher period and the New York Times datasets. $i$ and $j$ are the source and destination spaces of the mappings. Con and Lab are Conservative and Labour. Before and after are embeddings created for before 9/11 and after 9/11.} 
    \vspace*{-\baselineskip}
        \resizebox{0.45\textwidth}{!}{
        \begin{tabular}{c l c c}
        
    \toprule
    Dataset & Setting &  $cos(W^{ij}V^i_w, V^j_w)$ & $cos(W^{ji}W^{ij}V^i_w, V^i_w)$ \\
    \hline
     \multirow{2}{*}{UK} & $i= Con, j=Lab$ & 0.43 & 0.84 \\
      & $i= Lab, j=Con$ & 0.43 & 0.85\\
      \hline
     \multirow{2}{*}{NY Times} & $i= before, j=after$ & 0.49 & 0.85 \\
     & $i= after, j=before$ & 0.48 & 0.87 \\
    \bottomrule
    
    \end{tabular}
    }
\end{table}

\subsubsection{Parameter analysis}
\begin{figure}[t]
\centering
\includegraphics[width=\columnwidth]{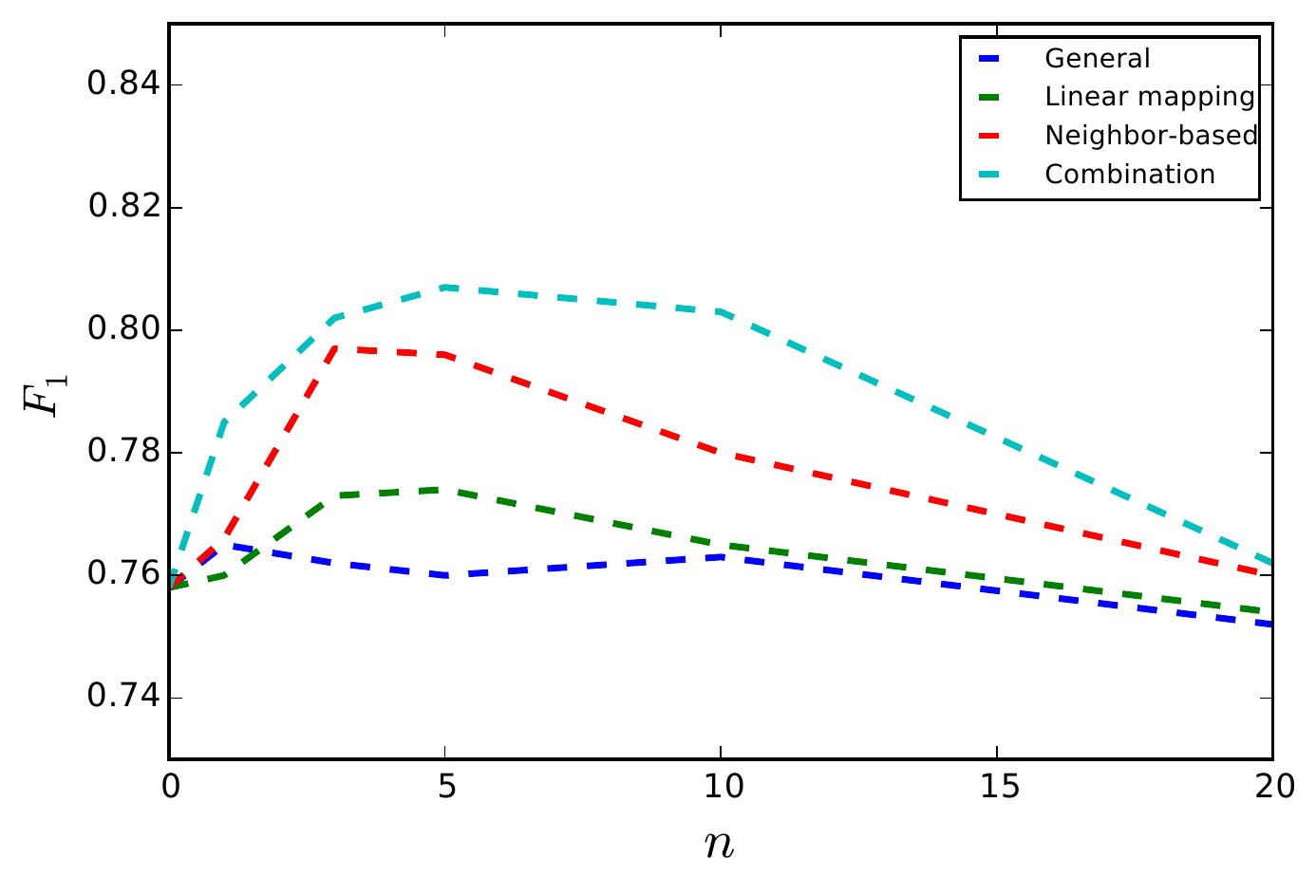}
\vspace*{-2.5\baselineskip}
\caption{The effect of the number of expansion words on the $F_1$ score of speech to party classification using different word stability measures. $n$ is the number of expansion words.}
\label{figure-param}
\end{figure}

In this section, we analyze the effect of the number of expansion words on the effectiveness of word stability approaches in the document classification task. Figure \ref{figure-param} shows the $F_1$ scores achieved using different methods based on different number of expansion words. For $1<n<3$, the $F_1$ score for all approaches is increased by increasing $n$. Moreover, for the Combination method even adding 5 expansion words boosts the performance of the classifier. This shows that adding more unstable words to documents can help in discriminating documents belonging to different parties. For $n>5$, adding more words decreases the performance of classifiers for all methods and the performance of all approaches are almost the same for $n=20$. This result indicates that by adding more than a certain number of words, the expanded documents become more and more similar, regardless of the measure used.

\section{Conclusion}
\label{conclusion}
We introduced a general framework for computing semantic shifts by using word embeddings trained on corpora that (are presumed to) represent specific viewpoints. We proposed several methods that compare words across these vector spaces--with their different dimensionality structures--and have demonstrated how these algorithms capture valuable changes in word meaning. We evaluated the results on political speeches and media reports. In doing so, we have shown that the techniques developed here adequately detect words that exhibit ideologically or chronologically diverging senses, \textit{and} can be applied to different types of discourse. We showed that semantic shifts not only occur over time, but also across viewpoints.

Our results demonstrated that the proposed word stability measures contribute to other tasks such as contrastive viewpoint summarization, which generates summaries that explicate the diverging viewpoints, and document classification. 
Moreover, we showed that temporal laws of change also apply to other dimensions. Our results demonstrated that the law of conformity strongly predicts the (in)stability of words, while the law of innovation only has a minimal effect. This indicates that the meaning of frequent words do not shift across viewpoints, while even the meaning of words with low polysemy values can shift. Furthermore, we proposed another law for semantic shifts which implies that more concrete words are insensitive to the viewpoint of speaker.

To the best of our knowledge, this paper is the first attempt to detect semantic shifts across viewpoints. We hope that the created datasets and proposed approaches will be beneficial to future research in this area. 

The estimated stability measures can be useful in various applications. As shown in this research, it can be used for summarizing diverging viewpoints and document classification. The generated summaries can be used in exploratory search scenario to uncover diverging aspects of a given topic. In this paper, we only focused on detecting shifts in political and media discourse, but our approaches are applicable in any other kind of discourse such as different groups in social media. 
 
Future work will focus on broadening the set of applications, by, for example, examining how our approach contributes to controversy detection and locating people in the ``filter bubble". If the language use of the specific group exhibits radically divergent word meanings, then they might be in the filter bubble and word stability can be used to quantify this. 

\mypar{Acknowledgements}
This research was supported by the Netherlands Organization for Scientific Research (ExPoSe project, NWO CI \# 314.99.108; DiLiPaD project, NWO Digging into Data \linebreak \# 600.006.014), Nederlab (340-6148-t1-6), and by the European Community's Seventh Framework Program (FP7/2007-2013) under grant agreement ENVRI, number 283465.

\bibliographystyle{abbrvnat}
\bibliography{ref}

\end{document}